\documentclass[11pt]{article}
\usepackage{UF_FRED_paper_style}
\usepackage[utf8]{inputenc} % allow utf-8 input
\usepackage[T1]{fontenc}    % use 8-bit T1 fonts
\usepackage{url}            % simple URL typesetting
\usepackage{booktabs}       % professional-quality tables
\usepackage{amsfonts}       % blackboard math symbols
\usepackage{nicefrac}       % compact symbols for 1/2, etc.
\usepackage{microtype}      % microtypography
\usepackage{amsmath}
\usepackage{caption}

\newcommand{\Lim}[1]{\raisebox{0.5ex}{{$\displaystyle \lim_{#1}\;$}}}
\usepackage[              
    hyperindex=true,                
    colorlinks=true,                
    breaklinks=true,                
    urlcolor= black,                
    linkcolor= blue,                
    bookmarks=true,                 
    bookmarksopen=false,
    filecolor=black,
    citecolor=blue]{hyperref} 

\usepackage{algorithm}
\usepackage{algpseudocode}
\usepackage{pifont}
\usepackage{algpseudocode,algorithm,algorithmicx}
\usepackage{nccmath}
\usepackage{booktabs}
\usepackage{caption}
\usepackage{siunitx}
\usepackage{adjustbox}
\usepackage{etoolbox}
\usepackage{listings}
\usepackage{color}
\usepackage{xcolor}
\usepackage{graphicx}
\usepackage{amsmath,amsthm,amssymb}

\renewrobustcmd{\bfseries}{\fontseries{b}\selectfont}
\renewrobustcmd{\boldmath}{}
\newrobustcmd{\B}{\bfseries}

\newtheorem{definition}{Definition}
\newtheorem{prop}{Proposition}

\title{Quantile Regularization: Towards Implicit Calibration of Regression Models
}

\author{Saiteja Utpala\\% Name author
    \href{sait@cse.iitk.ac.in}{\texttt{sait@cse.iitk.ac.in}} %% Email author 1 
\and Piyush Rai\\% Name author
    \href{piyush@cse.iitk.ac.in}{\texttt{piyush@cse.iitk.ac.in}} %%
    }
    
\date{CSE Department, IIT Kanpur}
\begin{document}
% %%%%%%%%%%%%%%%%%%%%%%%%%%%%%%%%%%%%%%%%%%%%%%%%%%%%%%%%%%
% %%%%%%%%%%%%%%%%%%%%%%%%%%%%%%%%%%%%%%%%%%%%%%%%%%%%%%%%%%
% ABSTRACT
% %%%%%%%%%%%%%%%%%%%%%%%%%%%%%%%%%%%%%%%%%%%%%%%%%%%%%%%%%%
% %%%%%%%%%%%%%%%%%%%%%%%%%%%%%%%%%%%%%%%%%%%%%%%%%%%%%%%%%%
{\setstretch{.8}
\maketitle
% %%%%%%%%%%%%%%%%%%
\begin{abstract}
 Recent works have shown that most deep learning models are often poorly calibrated, i.e., they may produce overconfident predictions that are wrong. It is therefore desirable to have models that produce predictive uncertainty estimates that are \emph{reliable}. Several approaches have been proposed recently to calibrate classification models. However, there is relatively little work on calibrating regression models. We present a method for calibrating regression models based on a novel quantile regularizer defined as the \emph{cumulative KL divergence} between two CDFs. Unlike most of the existing approaches for calibrating regression models, which are based on \emph{post-hoc} processing of the model's output and require an additional dataset, our method is trainable in an end-to-end fashion without requiring an additional dataset. The proposed regularizer can be used with any training objective for regression. We also show that post-hoc calibration methods like Isotonic Calibration sometimes compound miscalibration whereas our method provides consistently better calibrations. We provide empirical results demonstrating that the proposed quantile regularizer significantly improves calibration for regression models trained using approaches, such as Dropout VI and Deep Ensembles.
 
 %Calibration is notion that enables properly assessing the quality of uncertainties. Notion of calibration for classification has been well studied. However, there is sparse literature on Regression Calibration.Most of proposed methods are post hoc, where you need separate calibration dataset . We propose Quantile Regularization, a novel trainable calibration loss function that can be made part of training objective for regression networks,to achieve implicit calibration where you don't require additional dataset. We also show that posthoc calibration methods like Isotonic Calibration sometimes compound miscalibration.Finally, We verify that model probabilities are better calibrated when trained with quantile regularizer by using Dropout VI, Deep Ensembles as baselines

% END CONTENT ABS------------------------------------------
\noindent

\end{abstract}
}

% %%%%%%%%%%%%%%%%%%%%%%%%%%%%%%%%%%%%%%%%%%%%%%%%%%%%%%%%%%
% %%%%%%%%%%%%%%%%%%%%%%%%%%%%%%%%%%%%%%%%%%%%%%%%%%%%%%%%%%
% BODY OF THE DOCUMENT
% %%%%%%%%%%%%%%%%%%%%%%%%%%%%%%%%%%%%%%%%%%%%%%%%%%%%%%%%%%
% %%%%%%%%%%%%%%%%%%%%%%%%%%%%%%%%%%%%%%%%%%%%%%%%%%%%%%%%%%

% --------------------
\section{Introduction}

Calibration is a measure of evaluating how well a model's confidence in its prediction matches with the correctness of these predictions. For example, a binary classifier will be considered perfectly calibrated if among all predictions with probability score 0.9, 90\% of the predictions should be correct~\cite{guo2017calibration}. Likewise, consider a Bayesian regression model that produces credible intervals. In this setting, the model will be considered perfectly calibrated if the 90\% credible interval contains 90\% of the test points~\cite{kuleshov2018accurate}. Unfortunately, modern deep neural networks are known to be poorly calibrated~\cite{guo2017calibration}.

While there has been a significant amount of recent work on calibrating classification models~\cite{guo2017calibration,kumar2018trainable}, relatively little work exists on calibrating regression models. Recently, ~\cite{kuleshov2018accurate} proposed a post-hoc method for calibrating regression models. Their approach is inspired by Platt scaling~\cite{platt1999probabilistic}, commonly used for calibrating classification models. However, post-hoc methods like~\cite{kuleshov2018accurate} rely on the availability of large quantities of labeled i.i.d. data that is needed to achieve well-calibrated models.

In this work, we introduce \emph{quantile regularization}, a method that can be trained in an end-to-end manner unlike the post-hoc calibration methods that require large quantities of labeled data. The regularizer we proposed is defined as the \emph{cumulative KL divergence} between two CDFs. Moreover, our method has a very general applicability as it can be used in any regression model that produces a predictive mean and predictive variance, by augmenting its training objective with the proposed regularizer.

%Calibration is one of measures for evaluating uncertainty i.e., probabilistic predictions, just as RMSE/Accuracy are used to evaluate point predictions. Assume that we have Model that outputs vector of  real scores for multiclass classification.We can transform it such that model outputs probabilities using normalization function like softmax. But just because we are predicting normalized scores which lie in $[0,1]$ and sum up to $1$, mean they actually output \emph{probabilities}? If not, can we differentiate models that are actually predicting \emph{probabilities} not just normalized scores.Calibration is notion that is precisely developed for quantifying this. In broad Sense, Calibration makes sure that predicted normalized scores actually  approach empirical proportions.

Before describing our approach, we first provide a brief overview of calibration approaches proposed for classification and regression models.

\subsection{Classification Calibration}
\label{sec:classcalib}
The notion of calibration was originally first considered in meteorology literature \cite{brier1950verification,murphy1972scalar,gneiting2007strictly}  and saw one of its first prominent usage used in the machine learning literature by \cite{platt1999probabilistic} in context of support vector machines (SVM) in order to obtain probabilistic predictions from SVMs which are non-probabilistic models. There has been renewed interested in calibration, especially for classification models, after \cite{guo2017calibration} showed that modern classification networks are not well-calibrated.  

Currently there are three main notions of calibration in case of classification \cite{kumar2019verified,vaicenavicius2019evaluating,kull2019beyond} and these are listed below. For the rest of this section. assume $X,Y$ to be random variables on spaces $\mathcal{X}$ and $\mathcal{Y} = \{1,2,..K\}$, $\mathbb{P}$ to be their true joint distribution, and $g$ to be the model that outputs a probability distribution on $\mathcal{Y}$. Therefore, we can represent the model as $g: \mathcal{X} \rightarrow (\mathcal{Y} \rightarrow [0,1])$. The three notions are as follows:

\begin{enumerate}
    \item Top-Label calibration : $\mathbb{P}[Y = \arg \max  g(X)  |$  $\max g(X)] =  \max g(X), \,\, \forall g(X)$. It says that among all instances that the model predicts the most probable class with confidence say $p$, the proportion of instances that are actually of predicted class should be $p$.
    
    \item Marginal calibration  : $\mathbb{P}[Y = y | g(X)\{y\}] = g(X)\{y\},$ $\forall y \,\, \forall g(X)$. It says that, among all instances that the model predicts class $k$ with confidence $p$, the proportion of actual instances from class $k$ should be $p$ .
    
    \item Joint calibration  : $\mathbb{P}[Y = y |g(X)] = g(X)\{y\}, $ $\forall y \,\, \forall g(X)$. It says that, among all instances for which a distribution $\hat{p}$ is predicted , probability that it belongs to class $k$ is actually $\hat{p}[Y=k]$.
\end{enumerate}

\begin{figure}
    \centering
    \includegraphics[width = 0.6\linewidth]{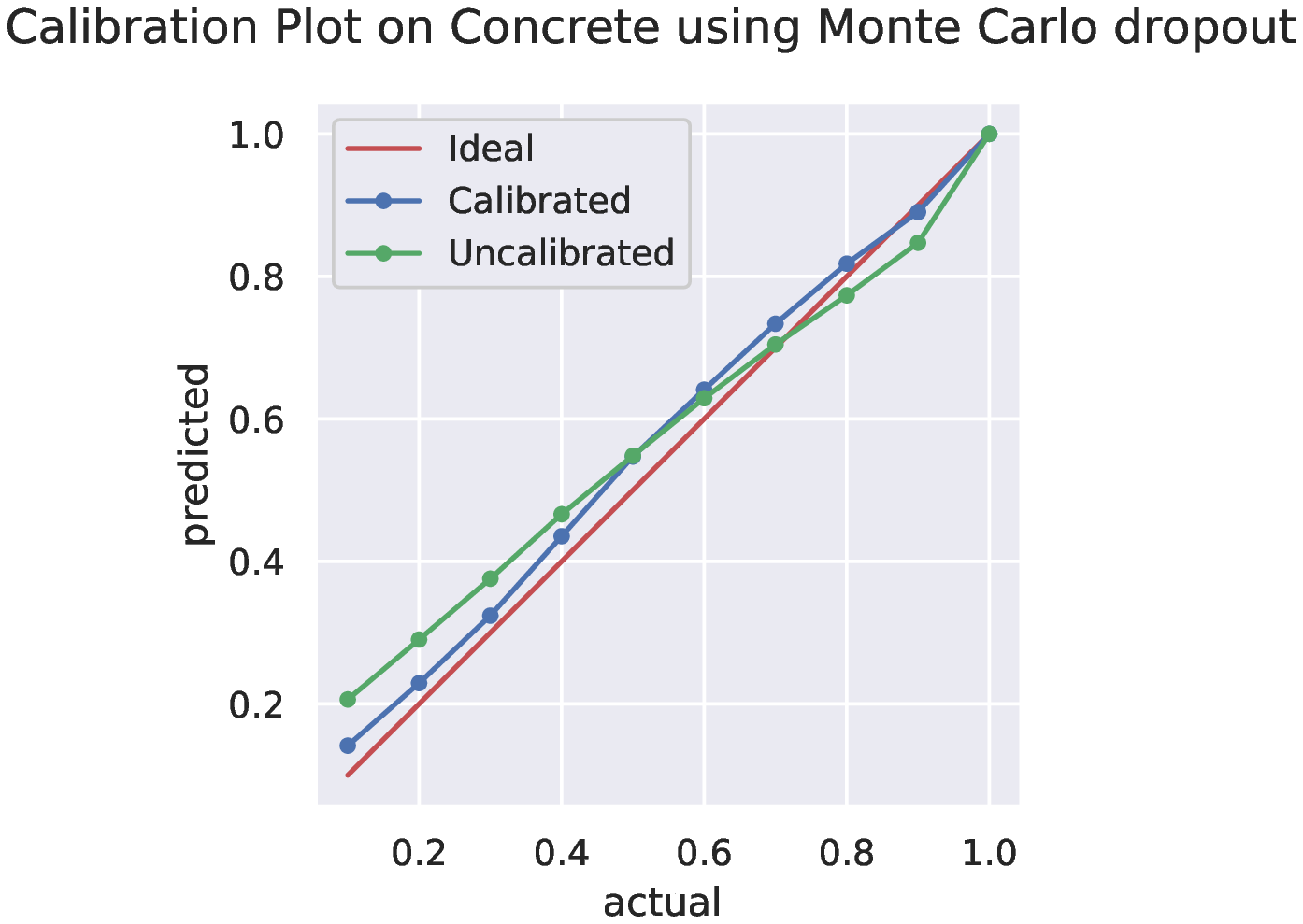}
    \caption{Calibration plot showing that Quantile Regularization makes calibration plot more close to the identity function (the ideal line) }
    \label{fig:my_label}
\end{figure}

Most calibration methods \cite{platt1999probabilistic,zadrozny2001obtaining,zadrozny2002transforming,guo2017calibration,kull2017beta,kull2019beyond} for classification models are \emph{post-hoc}, where they learn calibration mapping using an additional dataset to \emph{recalibrate} an already trained model. There has been recent work showing some of these popular post-hoc methods are either themselves miscalibrated or sample inefficient \cite{kumar2019verified} and they do not actually help the model output well-calibrated probabilities.

An alternative to post-hoc processing is to ensure that model outputs well-calibrated probabilities during training itself. These are implicit calibration methods. Such an approach does not require an additional dataset to learn the calibration mapping. While almost all post-hoc calibration mechanisms can be seen as density estimation methods, existing implicit calibration methods are of various types. Several heuristics like Mixup \cite{zhang2017mixup,thulasidasan2019mixup}  and Label Smoothing~\cite{szegedy2016rethinking, muller2019does} that were part of high performance deep networks for classification were later shown empirically to achieve calibration. \cite{maddox2019simple} show that their optimization method improves calibration. \cite{pereyra2017regularizing} found that penalizing high-confidence predictions acts as a regularizer. A more principled way of achieving calibration is by minimizing a loss function that is tailored for calibration \cite{kumar2018trainable}. This is somewhat similar in spirit to our proposed approach that does it for regression models.

\subsection{Regression Calibration}

There has been relatively less work on regression calibration. Among the early approaches, \cite{gneiting2007probabilistic} were the first to address this issue by proposing a framework for calibration. However, they do not provide any procedure to correct a mis-calibrated model. Recently, \cite{kuleshov2018accurate} proposed Quantile Calibration which intuitively says that the $p$ credible interval predicted by model should have target variable with probability $p$. They also propose a \emph{post-hoc} method based on isotonic regression \cite{fawcett2007pav} for recalibration which is a well-known recalibration technique for classification models. Recently, \cite{2019arXiv190506023S} proposed a much stronger notion of calibration called \emph{distributional calibration} which guarantees that among all instances whose predicted PDF has mean $\mu$ and standard deviation $\sigma$, the actual distribution of the target variable should have mean $\mu$ and standard deviation $\sigma$. This can be seen as the regression analog of \emph{joint calibration} for classification (Sec.~\ref{sec:classcalib}) . They too propose \emph{post-hoc} recalibration method based on Gaussian processes. Among other work, \cite{keren2018calibrated}, consider a different setting where neural networks for classification are used for regression problems and showed that temperature scaling \cite{hinton2015distilling,guo2017calibration} and their proposed method based on empirical prediction intervals improves calibration. Again, these are \emph{post-hoc} methods.

\subsection{Quantitle Calibration and Isotonic Regression}

The notion of calibration that we consider in this work is quantile calibration. Isotonic Regression is currently used for quantile calibration~\cite{kuleshov2018accurate}. However, isotonic regression has the following disadvantages

\begin{enumerate}
    \item It is a powerful nonparametric method that has tendency of overfitting, so much so that it perfectly passes through the datapoints if the datapoints already satisfy monotonicity constraint.
    
    \item Using an isotonic calibration mapping will result in a non-smooth and piecewise linear calibrated CDF. Consequently, the calibrated PDF is discontinuous. 
    
    \item It is a post-hoc method and ideally requires an additional dataset to learn the calibration mapping. 
\end{enumerate}

Considering these shortcomings, we propose an end-to-end trainable loss function for quantile calibration. Our approach leverages a novel regularizer that is defined as a cumulative KL divergence (KL divergence of two CDFs). With our approach, the smoothness of the PDF/CDF is maintained for well-calibrated probabilities. Moreover, our approach eliminates the need for a separate calibration dataset. To the best of our knowledge, this is the first trainable loss function for \emph{any} notion of calibration in regression setting.

The Rest of the paper is organized as follows: Section  \hyperref[sec:Background]{(2)} sets up the notation and background and presents the problem setting formally. In Section \hyperref[sec:QR]{(3)}, we present our proposed method. Section \hyperref[sec:Experiments]{(4)} discusses the experimental analysis. In Section \hyperref[sec:Future]{(5)}, we conclude and briefly discuss avenues for future work.

\section{Background and Definitions}
\label{sec:Background}

Throughout the paper, $X$ and $Y$ will denote random variables on spaces $\mathcal{X}$ and $\mathcal{Y} \subseteq \mathbb{R}$ with true distribution $\mathbb{P}$ and $(x_n , y_n)$ will denote i.i.d samples from this distribution.we assume that CDF's of random variables are invertible.

Any probabilistic regression model can be seen as \emph{conditional} CDF, which gives a distribution function on $\mathcal{Y}$ corresponding  to each instance from the input space $\mathcal{X}$. We represent the model as $M : \mathcal{X} \rightarrow (\mathcal{Y} \rightarrow [0,1]).$

Assume $F$ is distribution function predicted corresponding to the true distribution function $G$. Ideally we want to predict true distribution, i.e., $F=G$. This is equivalent of saying that $G(F^{-1}(p)) = p$ $\forall$ $p \in [0,1].$ Based on this,  \cite{gneiting2007probabilistic} propose the following definition

\begin{definition}[Complete Probabilistic Calibration]
Given a model $F: \mathcal{X} \rightarrow ( \mathcal{Y} \rightarrow [0,1] )$ and true underlying model  $G : \mathcal{X} \rightarrow (\mathcal{Y} \rightarrow [0,1])$ , the model $F$ is said to be probabilistically calibrated completely  iff for every sequence  $(x_{n})$

\begin{ceqn}
\begin{align}
  \Lim{N \rightarrow \infty} {\frac{1}{N} \sum_{n=1}^N G_n \circ F_n^{-1}(p) = p \,\,\,\,\, \forall p \in [0,1]} 
\end{align}
\end{ceqn}

\end{definition}

where $G_n= G(x_n)$ and $F_n = F(x_n)$

Since $G$ is unknown, \cite{kuleshov2018accurate} proposes the sufficient condition for above definition which is useful in practice.

\begin{definition}[Quantile Calibration]

Given a model $F : \mathcal{X} \rightarrow (\mathcal{Y} \rightarrow [0,1])$ and $X,Y$ jointly distributed as $\mathbb{P}$, the function $F$ is said to be Quantile Calibrated iff

\begin{ceqn}
\begin{align}
  \mathbb{P}\bigg[ [F(X)](Y) \leq p \bigg] = p \,\,\,\, \forall p \in [0,1]
\end{align}
\end{ceqn}

\end{definition}

The key to understanding above definition is the random variable under consideration $[F(X)](Y)$. Note that $[F(X)](Y)$ is cumulative density that the model predicts for random $X,Y$ whose underlying distribution is $\mathbb{P}$

The importance of such definition is that we get calibrated confidence/credible intervals, which is extremely critical for reliable uncertainty estimates. Its usefulness was demonstrated empirically in \cite{kuleshov2018accurate} who developed a post-hoc calibration method using the above notion of quantile calibration.

Existing calibration approaches can be divided into two types.

\begin{enumerate}
    \item Post-hoc Calibration: This approach recalibrates a pre-trained model using a separate calibration dataset by learning the \emph{canonical calibration mapping}~\cite{vaicenavicius2019evaluating}. 
    
    \item Implicit Calibration: This approach ensures that that model is calibrated while training itself without explicitly using a separate dataset.
\end{enumerate}

\subsection{Post-hoc calibration}

The objective of post-hoc calibration is to calibrate a miscalibrated model by learning a mapping $R : [0,1] \rightarrow [0,1]$ s.t $R \circ F$ is calibrated model. One such mapping can be obtained from definition of calibration itself. Setting $R(p) = \mathbb{P}\big[ [F(X)](Y) \leq p \big]$ makes $R \circ F$ a quantile calibrated model. Recently, \cite{vaicenavicius2019evaluating} refer to an analogous mapping in context of classification as \emph{canonical calibration mapping}. We will use same name to refer to it for our regression setting. 

\begin{prop}
For any Model $F: \mathcal{X} \rightarrow (\mathcal{Y} \rightarrow [0,1])$ and given the canonical calibration mapping $R(p) = \mathbb{P}\bigg[ [F(X)](Y) \leq p \bigg] $, $R \circ F$ is quantile calibrated
\end{prop}

The proof of this proposition can be found in the Appendix \hyperref[sec:Appendix]{(A1)} 

With this insight, and using the fact that mapping is monotonically increasing,~  \cite{kuleshov2018accurate} use isotonic regression to learn this mapping on the training dataset itself without using any separate dataset claiming that they do not overfit much. Given $\{(x_i ,y_i)\}_{i=1}^{n}$ , and assume that $x_1 \leq .... \leq x_n$, isotonic regression finds $\{\hat{y}_i\}_{i=1}^n$ by minimizing the following objective

\begin{ceqn}
\begin{align*}
 \textbf{y}^{*} = \arg\min_{\textbf{y} \in \mathbb{R}^n} \sum_{i=1}^n (y_i - \hat{y}_i)^2 \hspace{0.2cm} \text{subject to \,\,\,\,\,} \hat{y}_1 \leq .... \leq \hat{y}_n
\end{align*}
\end{ceqn}

In isotonic calibration~\cite{kuleshov2018accurate}, given training data $\{(x_i,y_i )\}_{i=1}^N$, the recalibration dataset is generated as $\{([F(x_i)](y_i) , \hat{P}(F(x_i)](y_i)) \}_{i=1}^n$ where $\hat{P}(p) = \frac{1}{n} \sum_{i=1}^n I[ [H(x_i)](y_i) \leq p]$. Then the isotonic calibration mapping is fit on this recalibration dataset. However, this approach can be prone to overfitting. One way to see why isotonic calibration can potentially overfit is that nature of recalibration dataset already satisfies the monotonicity constraint because $\hat{P}(p_1) \leq \hat{P}(p_2)$ if $p_1 \leq  p_2$. So, to minimize the loss, the calibration mapping \emph{passes} through $\{\hat{P}([F(x_i)](y_i))\}_{i=1}^n$ exactly. Also it is non-parametric methods that can overfit given less data. Therefore, \cite{kuleshov2018accurate} used training data itself in order to have plenty of data to learn the calibration mapping. Therefore, to recalibrate a pre-trained model you would need training data with which you would have trained the model. Another Disadvantage is that the isotonic mapping is a piecewise linear monotonic function, with which we have to compose our predicted CDF during test time. This results in non-smooth CDFs, which may not be desirable.

\subsection{Implicit Calibration}

In contrast to post-hoc calibration, implicit calibration ensures that the model is well-calibrated by having a strong inductive bias towards model parameters that yield well-calibrated predictions. Our approach can seen as regression analog of \cite{kumar2018trainable} where they designed a trainable loss function for classification by kernalizing the calibration error and  \cite{pereyra2017regularizing} where they minimize the entropy of softmax outputs.

\section{Quantile Regularization}
\label{sec:QR}
Recall that, in quantile calibration, we want $\mathbb{P}\big[ [F(X)](Y) \leq p \big] = p, \forall  p \in [0,1]$. Note that, both the right and the left hand sides can be seen as CDF of some random variables. Let $R(p) = \mathbb{P}\big[ [F(X)](Y) \leq p \big]$ and $S(p)=p$. Here $R$ can be seen as the the CDF of $[F(X)](Y)$ while $S$ can be seen as CDF of Uniform[0,1]. So quantile calibration essentially wants the two CDFs to be equal. This is equivalent to saying that, for perfectly calibrated quantile model, we have that $[F[X]](Y)$ is the Uniform[0,1] distribution. Our approach is based on this equivalence. Essentially, we penalize model if the r.v. $[F[X]](Y)$ deviates from Uniform[0,1]. This property can be used to design a calibration metric that can be trained with our loss function, yielding a well-calibrated model while training itself.

One possible divergence metric that one could use is the KL divergence. The KL divergence between a distribution and the uniform distribution is equal to differential entropy. This method will result in very interpretable way of getting calibration that is minimizing differentiable entropy of $(F[X])[Y]$. However, in practice, this would require using the Beta kernel \cite{chen1999beta} for density estimation and computing the entropy. Therefore, we use other divergences that can result in loss functions that are simpler to train.

\subsection{Cumulative KL divergence}

Cumulative KL divergence (CKL) \cite{baratpour2012testing} is based on cumulative residual entropy (CKL) \cite{rao2004cumulative}. We derive analytically closed-form expression for CKL  between a distribution with support on $[0,1]$ and Uniform[0,1], and use this divergence for our calibration method.

\begin{definition}[Cumulative Residual Entropy]
Let $S$ be non negative r.v with CDF $F_{S}$ and $\overline{F}_{S} = 1- F_{S}$ be survival function. Then the cumulative residual entropy is defined as 

\begin{ceqn}
\begin{align*}
 \epsilon(S) = - \int_0^{\infty} \overline{F}_{S}(s) \log \overline{F}_{S}(s) \, ds
\end{align*}
\end{ceqn}

\end{definition}

\begin{definition}[Cumulative KL divergence]
Let $S,T$ be non-negative r.v with CDF $F_{S},G_{T}$ and $\overline{F}_{S} = 1-F_{S} ,\overline{G}_{T} = 1-G_{T}$ be corresponding survival functions . Then the cumulative KL divergence between $S$ and $T$ is defined as 

\begin{ceqn}
\begin{align*}
    \text{CKL}(F_{S} || G_{T}) =  \int_0^{\infty} \overline{F}_{S}(x) \ln \frac{\overline{F}_{S}(x)}{\overline{G}_{T}(x)} dx - \mathbb{E}[S] + \mathbb{E}[T] 
\end{align*}
\end{ceqn}

\end{definition}

The cumulative KL divergence has similar properties as the standard KL divergence. In particular, $CKL(F_{S}||G_{T}) \geq 0$ for any CDF's $F_{S},G_{T}$, and $CKL(F_{S}||G_{T})=0$ \text{iff} $F_{S}=G_{T}$

\begin{prop}
Consider random variable $S$ with CDF $F_{S}$ with support $[0,1]$ and let $T\sim \text{Uniform[0,1]}$ with CDF $G_{T}$ then CKL in terms of residual entropy is as follows

\begin{ceqn}
\begin{align}
   \text{CKL}(F_{S} || G_{T}) =  - \epsilon(S) + \mathbb{E}[ (1-S) \ln(1-S)] + 0.5
\end{align}
\end{ceqn}
\end{prop}

Proof of the above proposition can be found in the Appendix  \hyperref[sec:Appendix]{.A1}

\begin{prop}
Given $\{s_k\}_{k=1}^{n} \overset{iid}{\sim} F_{S}$, let $s_{(1)} \leq s_{(2)} .... \leq s_{(n)}$ denote ordered samples, then the following is a consistent estimator of above expression
 
\begin{equation}
\label{eq:ckl_esp}
\begin{aligned}
\overline{CKL}(F_S ||G_T) ={} & \sum_{i=1}^{n-1} \frac{n-i}{n} \Big (\ln \frac{n-i}{n}  \Big)  s_{(i+1)} - s_{(i)} + \frac{1}{n}\sum_{i=1}^n (1- s_i) \ln (1-s_i) +0.5 \\
\end{aligned}
\end{equation}
\end{prop}

Proof of the above proposition can be found in the Appendix  \hyperref[sec:Appendix]{.A1}

\subsection{Calibration loss function}

In our case, the random variable is $[F(X)](Y)$ where $F$ is the model. Given i.i.d. samples $(\textbf{x}_k,y_k)$ in the training data, we need to generate samples $[F(\textbf{x}_k)](y_k)$ to compute the expression given in Eq.~\ref{eq:ckl_esp}.

 Note that, we want to make this part of the training procedure to achieve implicit calibration. However, we are faced with a challenge here. In particular, we need  \emph{ordered} samples to compute the first summation in Eq.~\ref{eq:ckl_esp} whereas sorting is not a differentiable operation. There are many differentiable approximations to sorting operation.We use NeuralSort \cite{grover2019stochastic} for its simplicity in our experiments. The algorithm for computing the loss function is summarized below.

\newcommand*\MODEL{\textsc{model}}
\newcommand*\SORT{\textsc{diffsort}}
\newcommand*\NLL{\textsc{nll}}
\newcommand*\CL{\textsc{cl}}
\newcommand*\CE{\textsc{ce}}

\newcommand*\Let[2]{\State #1 $\gets$ #2}
\algrenewcommand\algorithmicrequire{\textbf{Precondition:}}
\algrenewcommand\algorithmicensure{\textbf{Postcondition:}}

\begin{algorithm}
  \caption{Quantile Regularization
    \label{alg:packed-dna-hamming}}
  \begin{algorithmic}[1]
    \Require{$(\textbf{x}_k,y_k)$ are i.i.d training instances and $\mu_k,\sigma_k = \MODEL{}_{\textbf{w}}(\textbf{x}_k) $ and \SORT{} is any differentiable relaxation to sorting operation. }
    \Statex
    \Function{Calibration Loss function}{$\textbf{y} , \boldsymbol{\mu} , \boldsymbol{\sigma}$} \\
        \For{$k \gets 1 \textrm{ to } n$}
        
            \Let{$\Phi_k$}{$\Phi(\mu_k, \sigma_k)$} \Comment{$\Phi$: CDF}
             \Let{$c_k\,$}{$\, \Phi_k[y_k]$}
             \Let{$s_k \,$}{\SORT{}(c)[k]}
       
       \EndFor

      \Let{$a \,\,\,\,$}{$\sum_{i=1}^{n-1} \frac{n-i}{n} \big (\ln \frac{n-i}{n}\big)  s_{i+1} - s_{i}$}
      \Let{$b \,\,\,\,$}{$\frac{1}{n}\sum_{i=1}^n (1- s_i) \ln (1-s_i)$}
      \Let{$l\,\,\,\,\,$}{$a+b+0.5$} \\

      \State \Return{$l$}
    \EndFunction
  \end{algorithmic}
\end{algorithm}

 The overall loss function with quantile regularization is as  follows: Given training data $(\textbf{X}, \textbf{y})$, let $(\boldsymbol{\mu}_{\textbf{w}},\boldsymbol{\sigma}_{\textbf{w}}) = \MODEL{}_{\textbf{w}}(\textbf{X})$ ,   $\textbf{w}$ be parameters of the model, $\mathcal{NLL}(\textbf{y},\boldsymbol{\mu}_{\textbf{w}}, \boldsymbol{\sigma}_{\textbf{w}})$ be the negative log likelihood and $\mathcal{CL}(\textbf{y},\boldsymbol{\mu}_{\textbf{w}}, \boldsymbol{\sigma}_{\textbf{w}})$ be the calibrated loss computed by Algorithm 1.

\begin{ceqn}
\begin{align*}
  \mathcal{L}(\textbf{X},\textbf{y} ,\boldsymbol{\mu}_{\textbf{W}},\boldsymbol{\sigma}_{\textbf{W}})  = \mathcal{NLL}(\textbf{y},\boldsymbol{\mu}_{\textbf{w}},\boldsymbol{\sigma}_{\textbf{w}}) + \lambda \times   \mathcal{CL}(\textbf{y},\boldsymbol{\mu}_{\textbf{w}},\boldsymbol{\sigma}_{\textbf{w}})
\end{align*}
\end{ceqn}

\subsection{Sharpness with Calibrated Predictions}

Note that calibration is alone not sufficient for predictions to be accurate; sharpness is needed too. Our method can seen as naturally achieving both desiderata. While the usual negative log-likelihood (NLL) makes sure that the prediction are sharp, the quantile regularizer makes sure that those predictions are calibrated too, with $\lambda$ controlling strength of the regularization. As our experiments show, the RMSE and NLL scores do not worse much for even values as large as $\lambda = 20$.

\section{Experiments}
\label{sec:Experiments}

We evaluate our approach on various regression datasets in terms of the calibration error as well as other standard metrics, sich as root-mean-squared-error (RMSE) and negative log-likelihood (NLL). We experiment with two base models - MC Dropout~\cite{gal2016dropout} and \cite{lakshminarayanan2017simple} - by augmenting their objective functions with our proposed quantile regularizer.

\subsection{Metrics}

\subsubsection*{ $l_2$ Quantile Calibration Error} Given any model $F: \mathcal{X} \rightarrow (\mathcal{Y} \rightarrow [0,1])$, we define the $l_{2}$ calibration error as follows

\begin{ceqn}
\begin{align*}
    \mathcal{CE}(F) = \int_0^1 \big( \mathbb{P}\big[ [F(X)](Y) \leq p \big] - p \big)^2 dp
\end{align*}
\end{ceqn}

\begin{table}[]
    \centering
   
       \label{tab:first}
    \scalebox{0.8}{\begin{tabular}{ccccccc}
        \toprule
        Dataset & \multicolumn{6}{c}{HeteroScedastic MC Dropout model}                                                                          \\ \cmidrule{2-7} 
        & \multicolumn{2}{c}{Calib Error(\%)} & \multicolumn{2}{c}{RMSE} & \multicolumn{2}{c}{NLL} \\ \cmidrule{2-7} 
        & base              & QR       & base              & QR         & base      & QR   \\
        \midrule
       Air Foil  &  25.92 $\pm$ 5.40  & \B 19.00 $\pm$ 4.80  & \B 3.26 $\pm$ 0.06 & 3.75 $\pm$ 0.10 & \B 2.58 $\pm$ 0.03 & 2.71 $\pm$ 0.02  \\
        
       Boston Housing      & 44.99 $\pm$ 4.41 &  \B 42.59 $\pm$ 5.30 & 4.67 $\pm$ 0.06 & \B 4.65 $\pm$ 0.17 &  3.73 $\pm$ 0.32 &  \B 3.32 $\pm$ 0.10  \\

        Concrete Strength      & 58.75 $\pm$ 8.31 & \B 35.42 $\pm$ 6.32 & \B 8.61 $\pm$ 0.16 & 8.98 $\pm$ 0.14 &  3.74 $\pm$ 0.04 & \B 3.64 $\pm$ 0.02  \\

         Fish Toxicity      &4.22 $\pm$ 1.56 & \B 3.94 $\pm$ 0.78 & \B 0.92 $\pm$ 0.00  & 0.96 $\pm$ 0.01 & 1.25 $\pm$ 0.01 & \B 1.24 $\pm$ 0.01    \\
         
          Kin8nm       &  12.37 $\pm$ 0.61 &  \B 11.60 $\pm$ 1.12 &  \B 0.09 $\pm$ 0.00 & 0.10 $\pm$ 0.00 &  \B-0.87 $\pm$ 0.00 & -0.80 $\pm$ 0.01     \\

           Protein  Structure      & 5.49 $\pm$ 0.59 & \B 3.79 $\pm$ 0.82 &  \B 4.41 $\pm$ 0.06 &  \B 4.41 $\pm$ 0.08  &  2.82 $\pm$ 0.03  & \B 2.81 $\pm$ 0.01    \\ 
            Red Wine      & 8.86 $\pm$ 1.11 &  \B 6.78 $\pm$ 0.99  &  0.68 $\pm$ 0.00 & \B 0.66 $\pm$ 0.01 &  1.20 $\pm$ 0.03 &  \B 1.04 $\pm$ 0.01        \\ 
            White Wine    & 9.69 $\pm$ 1.96 & \B 7.63 $\pm$ 1.82 & 0.75 $\pm$ 0.01 &  \B 0.73 $\pm$ 0.00 & 1.20 $\pm$ 0.01 & \B 1.14 $\pm$ 0.01   \\ 
            
            Yacht  Hydrodynamics      & 55.21 $\pm$ 9.38 &  \B 40.08 $\pm$ 8.85 & \B 3.55 $\pm$ 0.63 & 5.50 $\pm$ 0.40 & \B 2.30 $\pm$ 0.12 & 2.77 $\pm$ 0.03   \\ 
            Year Prediction MSD      & 8.52 $\pm$ 2.42 &   \B 3.89 $\pm$ 3.56 &  \B 9.18 $\pm$ 0.12 & 9.20 $\pm$ 0.16 & 3.45 $\pm$ 0.03 & \B 3.42 $\pm$ 0.01     \\

        \bottomrule
    \end{tabular}}
     \caption{\emph{base} and \emph{QR}  stands for model trained without Quantile Regularization and with Quantile Regularization respectively. As we can see, the calibration error is reduced, all the while keeping RMSE/NLL close/better to/than the \emph{base} model. }
\end{table}

Let us choose $m$ equidistant points  $\{p_m\}_{m=1}^M $ in $(0,1]$ with $p_M=1$. Given a test set $\{x_n,y_n\}_{n=1}^N$ whose predictions are $F_n = F(x_n)$, the  $M$-bin estimator of above integral will give us the following metric used in \cite{kuleshov2018accurate}

\begin{ceqn}
\begin{align*}
    \mathcal{CE}(F,x,y,p) = \frac{1}{M} \sum_{i=1}^M \Big[  \sum_{j=1}^N \frac{1}{N}  \text{I} [  F_j(y_j) \leq p_i ] - p_i \Big]^2
\end{align*}
\end{ceqn}

\subsection{Models}

\subsubsection{Heteroscedastic MC Dropout}

We integrate our quantile regularizer with the heteroscedastic MC dropout approach \cite{gal2016uncertainty} where, for each instance, a neural network with Dropout predicts $(\mu,\sigma^2)$ and is trained with Gaussian likelihood . While testing, we enable dropout and  perform $T$ stochastic forward passes and set $\mu = \sum_{i=1}^T \mu_i$ and $\sigma^2 = \sum_{i=1}^{T} \sigma_i^2 + \frac{1}{T} \sum_{i=1}^T \Big( \sigma_i- \mu \Big)^2$ and posit $\text{Normal}(\mu,\sigma^2)$ as our prediction. Dropout rate is set to $0.25$ and we perform $T=10$ forward passes.

\subsubsection{Deep Ensembles :}

We also test our quantile regularizer method using deep ensembles \cite{lakshminarayanan2017simple} as they also provide uncertainty estimates. We fix the ensemble size = 5 where each network has Adversarial Training with $\epsilon = 0.01 * range[x_i]$ where $range[x_i]$ is range of input features along that dimension,  as suggested in the paper~\cite{lakshminarayanan2017simple}.

\subsection{Hyperparameters}

 We use the same hyperparamter settings, for all the models and all the datasets. In particular, we use a two hidden-layer network with 128 units and learning rate = 1e-2 with Adam Optimizer identical to \cite{kuleshov2018accurate} and batch size = 512, amd number of epochs = 100.

\subsection{UCI datasets }

We experiment with the following datasets (size-of-data,num-input-features): AirFoil (1503,6) , Boston Housing (506,13), Concrete Strength (1030,8), Fish Toxicity (908,7), Kin8nm (8192, 9), Protein Structure (45730, 10), Red Wine (1599, 12), White Wine (4898, 12), Yacht Hydrodynamics (308,6), and Year Prediction MSD (515345,91). The dataset sizes range from $308 $ to $515345$ and input feature dimensions range from $6$ to $91$. Every dataset, except Year Prediction MSD, is split into 5 splits whereas, for Year Prediction MSD, there is pre-defined single split where we train on 463715 points and test on 51630 points. Each experiment is repeated 5 times and averages are reported.

\begin{table}[]
    \centering
   
      \label{tab:second}
    \scalebox{0.8}{\begin{tabular}{ccccccc}
        \toprule
        Dataset & \multicolumn{6}{c}{Deep Ensembles}                                                                          \\ \cmidrule{2-7} 
        & \multicolumn{2}{c}{Calib Error(\%)} & \multicolumn{2}{c}{RMSE} & \multicolumn{2}{c}{NLL} \\ \cmidrule{2-7} 
        & base              & QR       & base              & QR         & base      & QR   \\
        \midrule
         Air Foil & 45.04 $\pm$ 0.86  & \B 30.63 $\pm$ 3.67 & \B 3.33 $\pm$ 0.04 & 3.46 $\pm$ 0.04 & 2.66 $\pm$ 0.01 & \B 2.65 $\pm$ 0.03    \\
        
       Boston Housing     & 8.19 $\pm$ 2.07 &  \B 5.89 $\pm$ 1.50 & \B 3.27 $\pm$ 0.06 & 3.45 $\pm$ 0.02 & \B 2.65 $\pm$ 0.06 & 2.73 $\pm$ 0.05 \\

        Concrete Strength   & 81.34 $\pm$ 4.61 & \B 65.48 $\pm$ 3.27 & 10.27 $\pm$ 0.15  & \B 9.76 $\pm$ 0.13  & 4.63 $\pm$ 0.23 & \B 4.28 $\pm$ 0.05   \\
        
         Fish Toxicity      &5.35 $\pm$  0.65 & \B 4.39 $\pm$ 1.23 & \B 0.90 $\pm$ 0.00 & \B 0.90 $\pm$ 0.01 & 1.41 $\pm$ 0.02 & \B 1.32 $\pm$ 0.02      \\
         
          Kin8nm       & \B 1.22 $\pm$ 0.30 & 1.54 $\pm$ 0.43 & \B 0.07 $\pm$ 0.00 & \B 0.07 $\pm$ 0.00 & \B -1.34 $\pm$ 0.00 & \B -1.34 $\pm$ 0.00     \\ 
          
           Protein Structure   & 5.09 $\pm$ 0.39 & \B 3.07 $\pm$ 0.20 & \B 4.14 $\pm$ 0.00 & 4.22 $\pm$ 0.02 & \B  2.67 $\pm$ 0.00 & 2.69 $\pm$ 0.00       \\ 
          
            Red Wine     &9.56 $\pm$ 1.21 &  \B 7.16 $\pm$ 1.15 & 0.66 $\pm$ 0.00 & \B 0.65 $\pm$ 0.00 & 1.33 $\pm$ 0.03 & \B 1.14 $\pm$ 0.03       \\ 
            White Wine    & 8.42 $\pm$ 0.55 & \B 8.06 $\pm$ 0.86 & 0.73 $\pm$ 0.00 & \B 0.73 $\pm$ 0.00 & 1.19 $\pm$ 0.01 & \B 1.15 $\pm$ 0.01    \\ 
            Yacht  Hydrodynamics      &84.38 $\pm$ 4.58 &  \B 54.23 $\pm$ 5.65 & \B 3.85 $\pm$ 0.22 & 4.69 $\pm$ 0.24 & 2.90 $\pm$ 0.18 & \B 2.30 $\pm$ 0.10    \\ 
            
            Year Prediction MSD      & 6.57 $\pm$ 1.39 & \B 2.41 $\pm$ 1.22 & \B  8.89 $\pm$ 0.09 & 8.96 $\pm$ 0.18 & \B 3.38 $\pm$ 0.02 & \B 3.38 $\pm$ 0.02   \\ 
        \bottomrule
    \end{tabular}}
     \caption{\emph{base} and \emph{QR}  stands for model trained without Quantile Regularization and with Quantile Regularization respectively. As we can see, the calibration error is reduced, all the while keeping RMSE/NLL close/better to/than the \emph{base} model. }
\end{table}

~

\textbf{HeteroScedastic MC Dropout Model}

\hyperref[tab:first]{Table 1} reports Calibration error, RMSE, NLL, and recalibration error when trained with and without quantile regularization. As shown in the table, the calibration error is smaller for the variant of MC Dropout model when model is trained with Quantile Regularization. In 7/10 cases, even the NLL is better. RMSE drops, if any, are almost negligible.

~
\textbf{Deep Ensembles}

\hyperref[tab:second]{Table 2} reports Calibration error, RMSE, NLL, and recalibration error with Deep Ensembles as underlying model. With Deep Ensembles as the base model, one can see that quantile regularization decreases calibration error in $9/10$ cases.

\textbf{Choice of the hyperparameter $\lambda$ }

\begin{figure}
    \centering
    \includegraphics[width = 0.3 \linewidth]{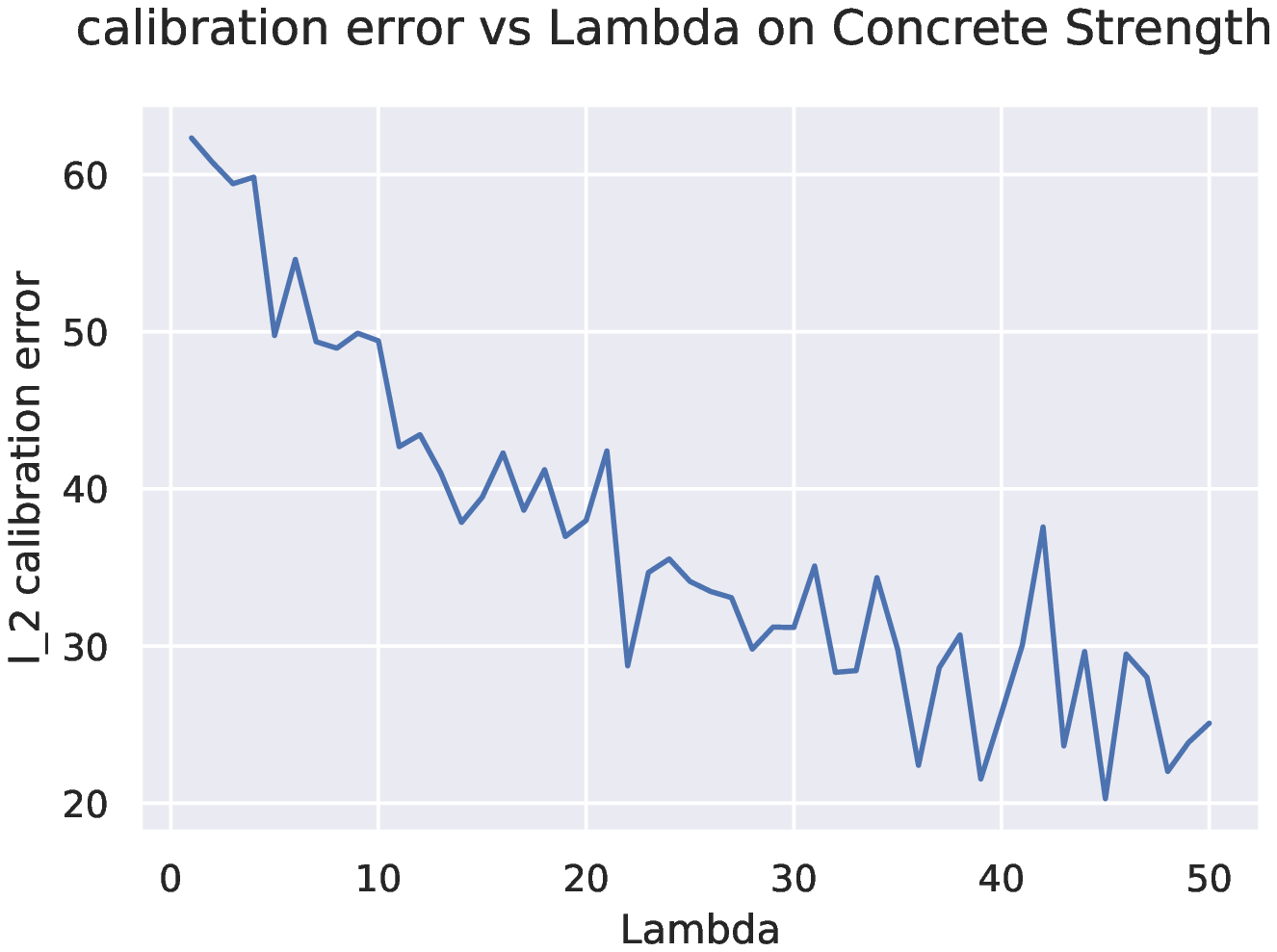}
    \includegraphics[width = 0.3 \linewidth]{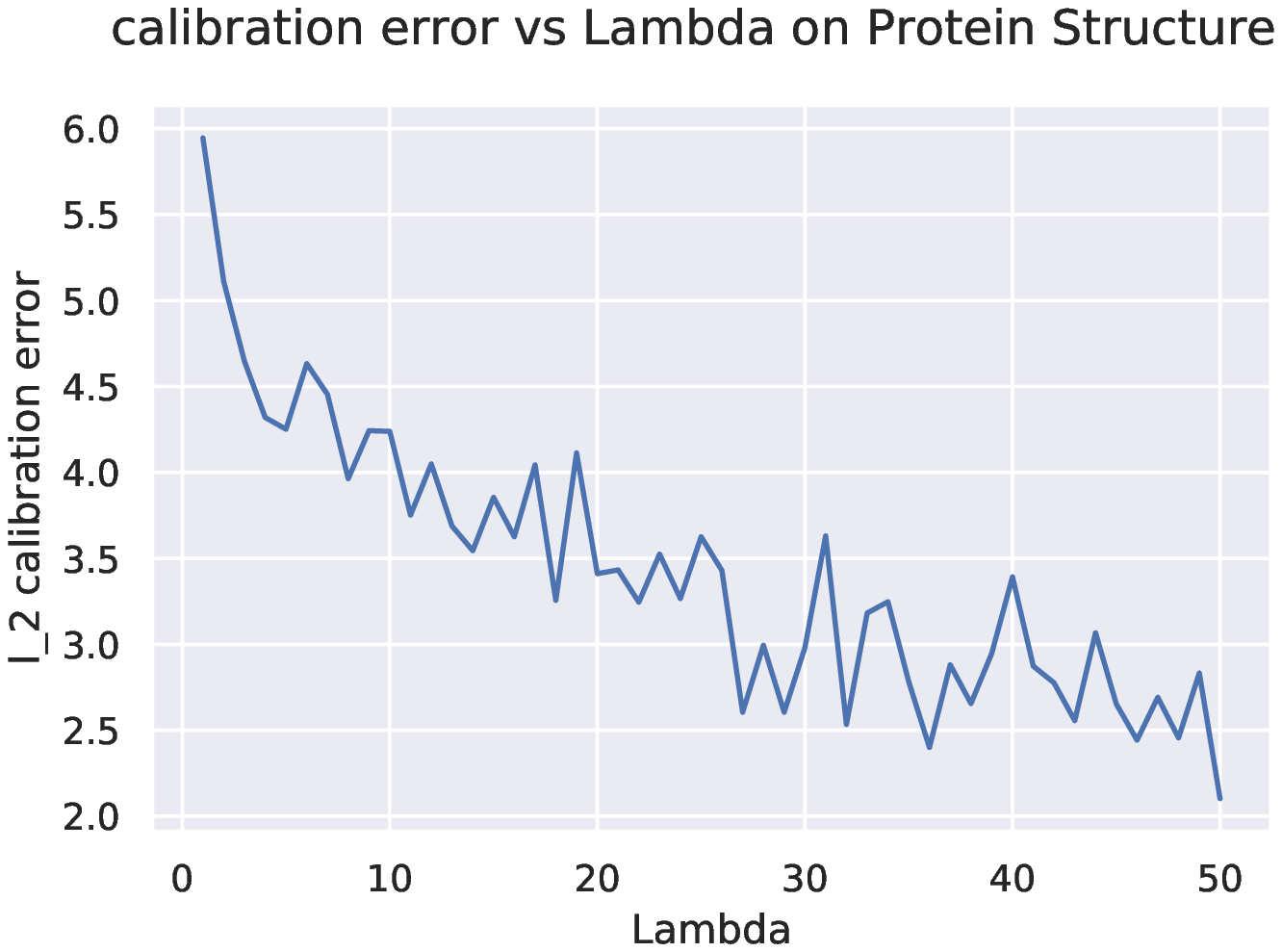}
    \includegraphics[width = 0.3 \linewidth]{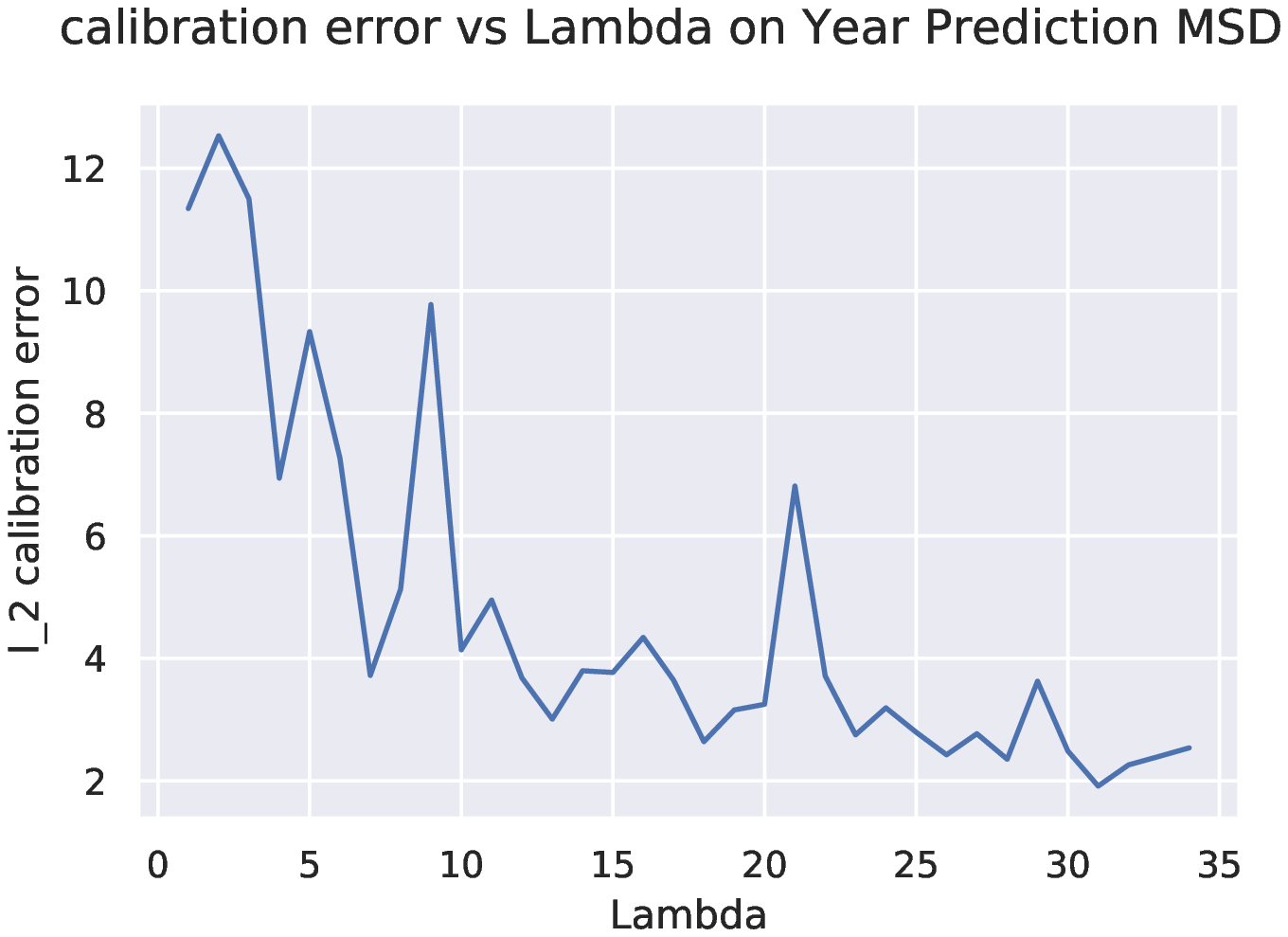}

    \includegraphics[width = 0.3 \linewidth]{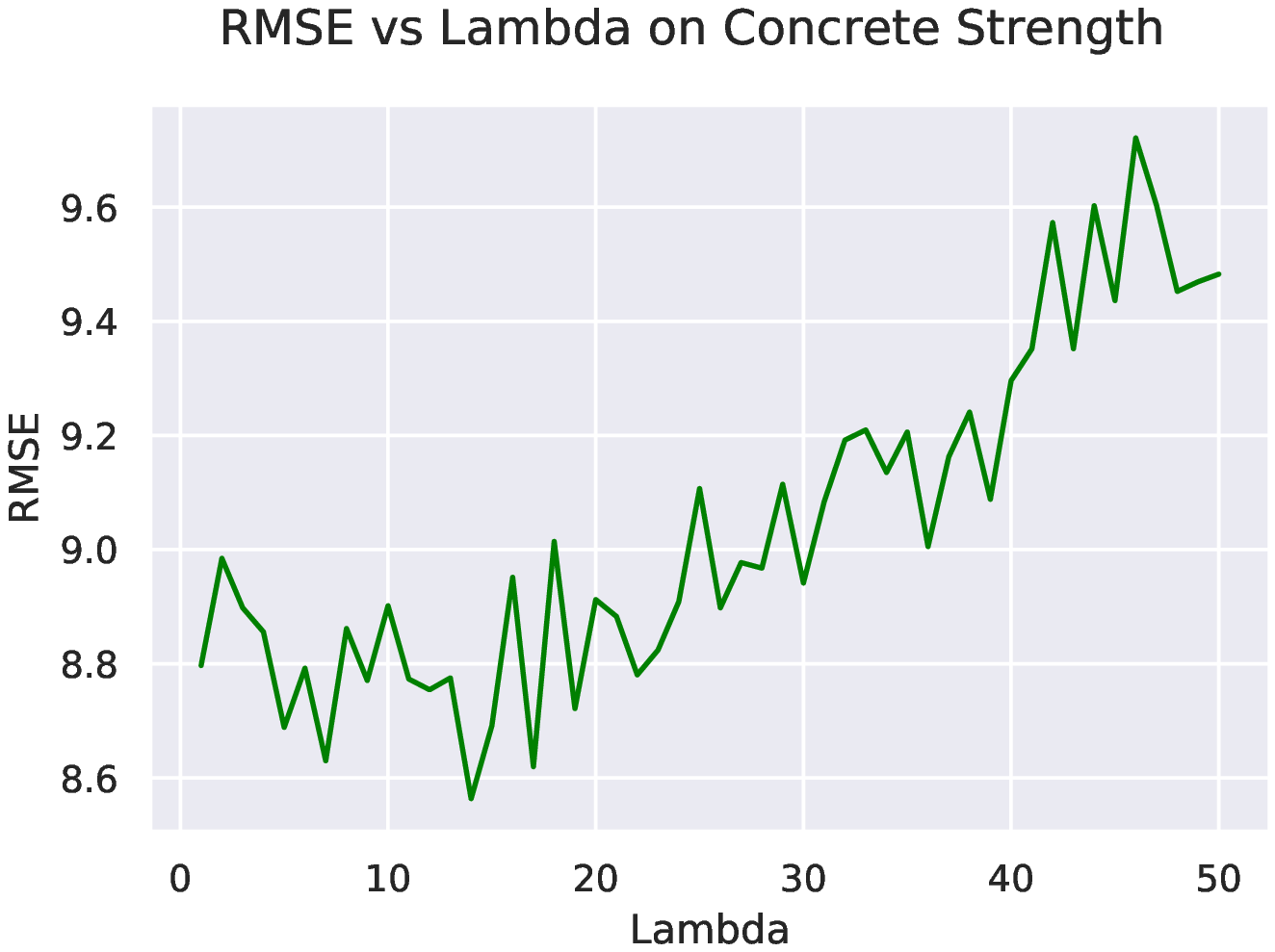}
    \includegraphics[width = 0.3 \linewidth]{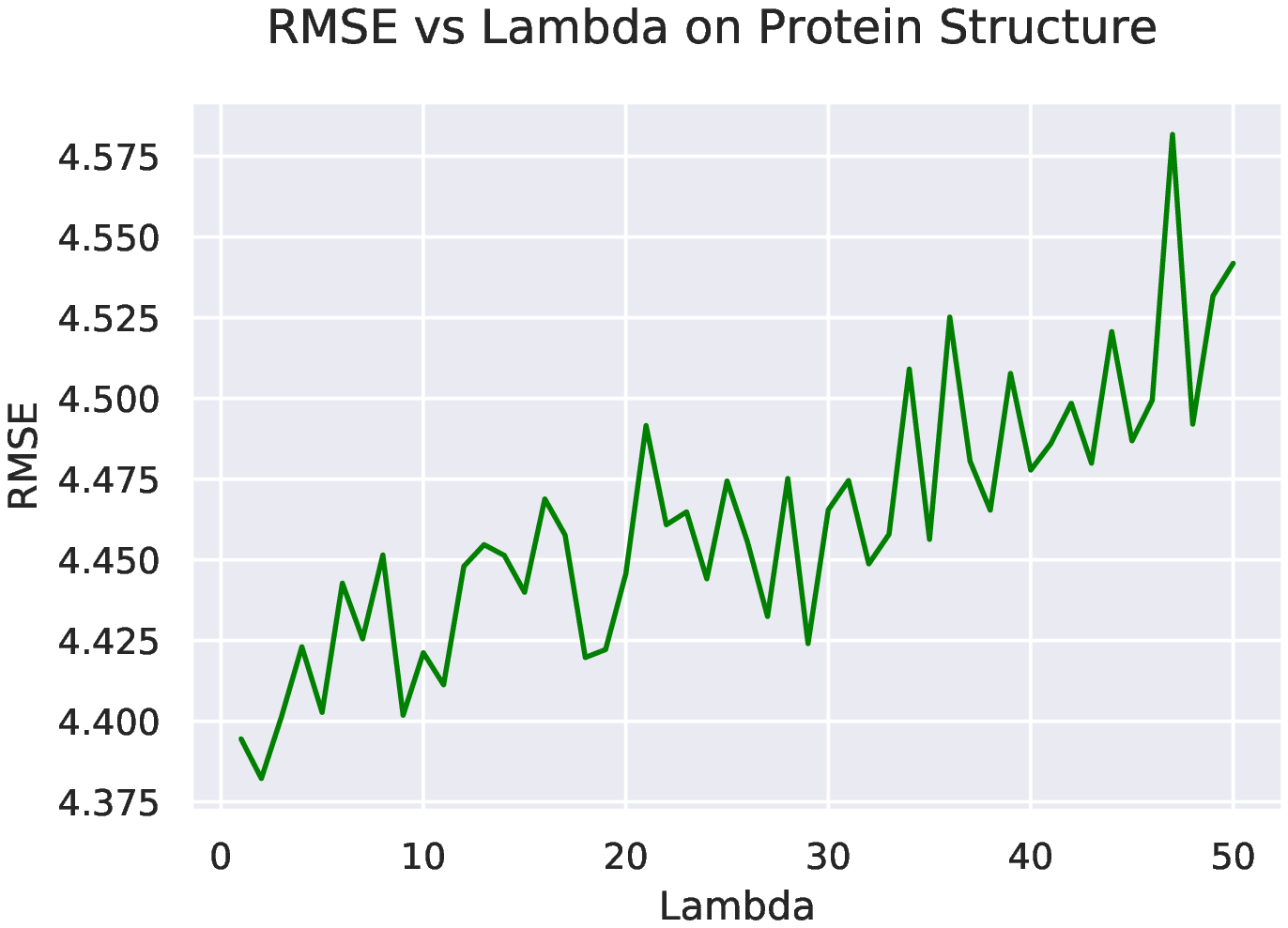}
    \includegraphics[width = 0.3 \linewidth]{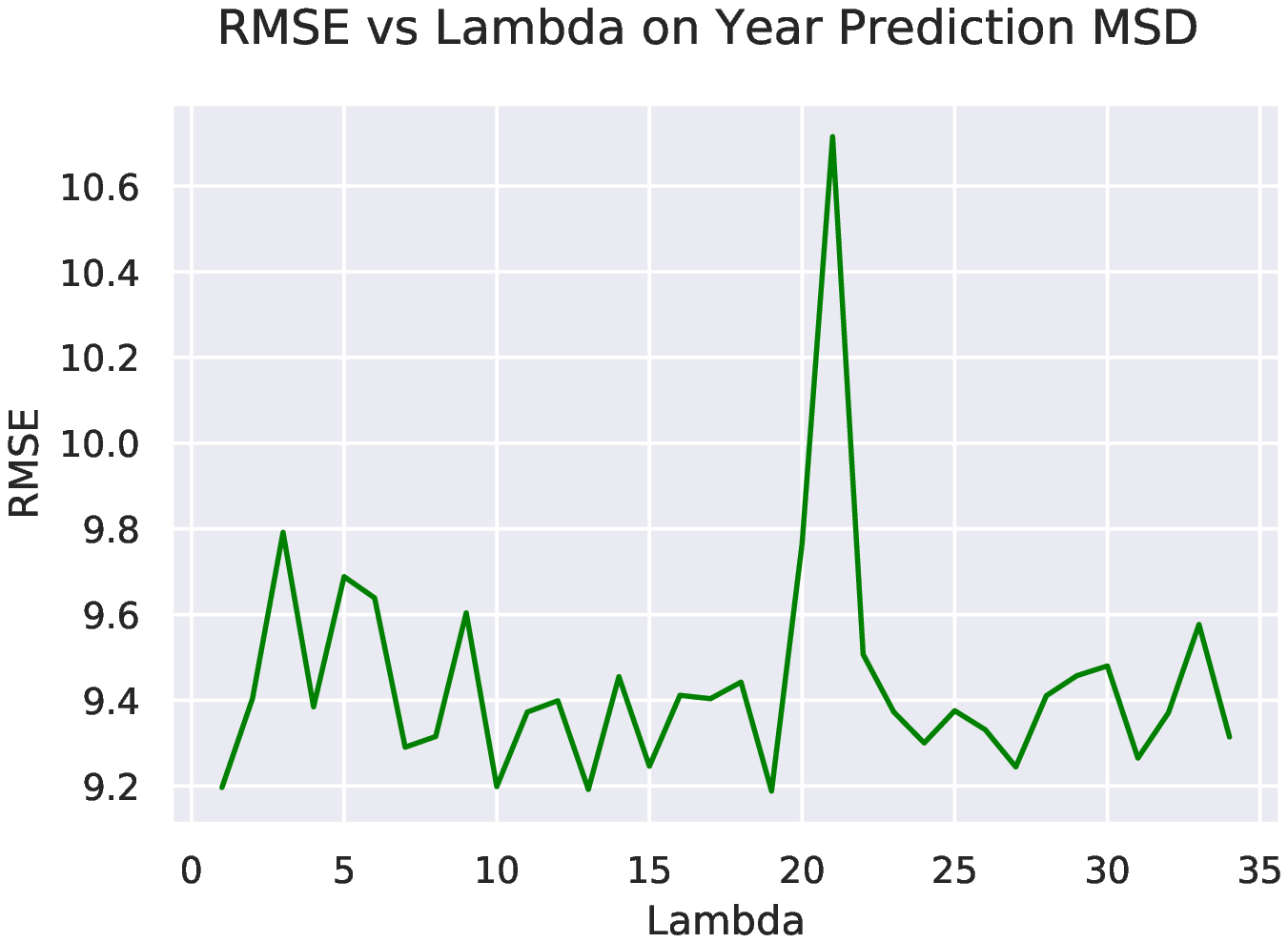}

    \includegraphics[width = 0.3 \linewidth]{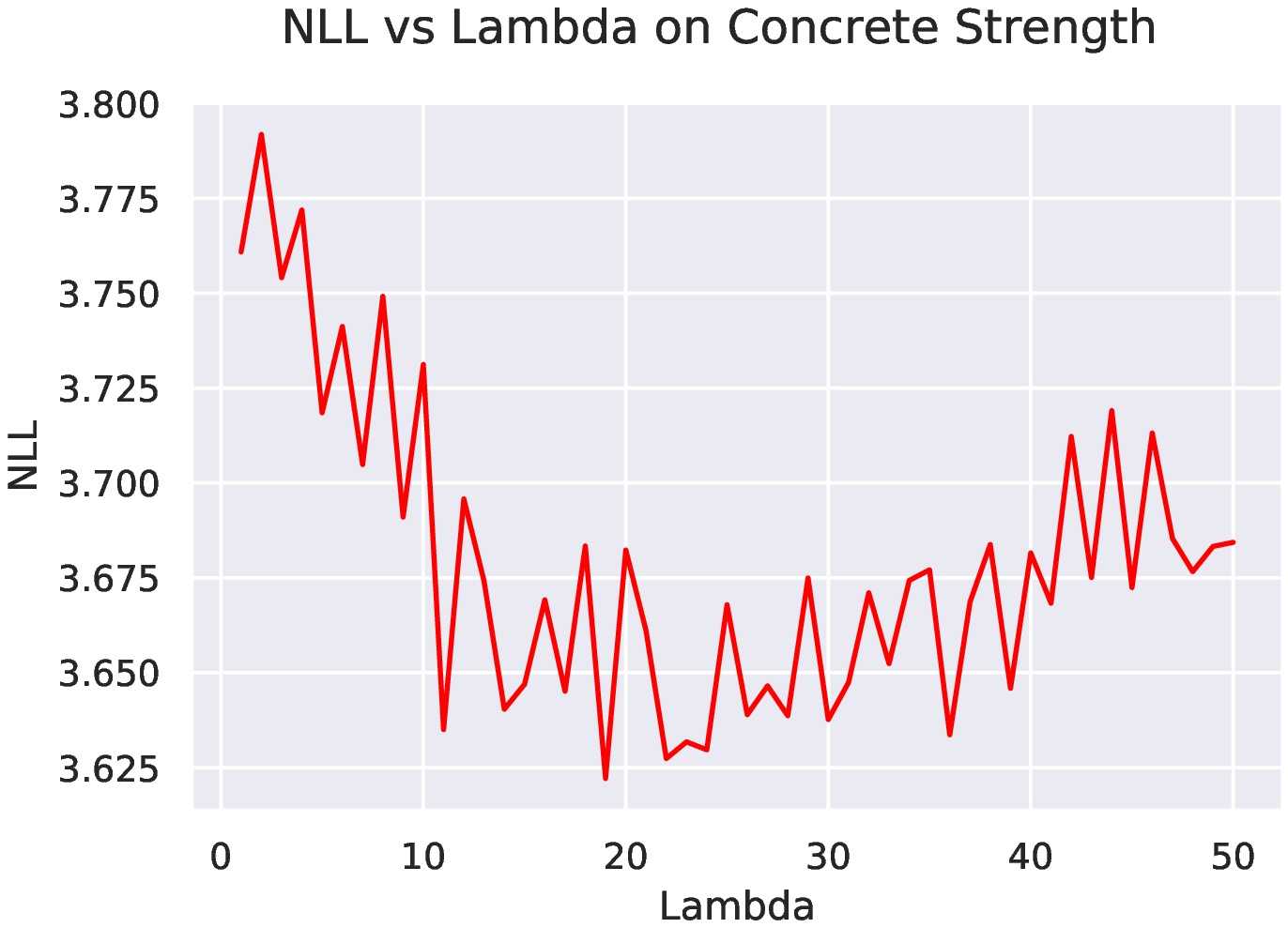}
    \includegraphics[width = 0.3 \linewidth]{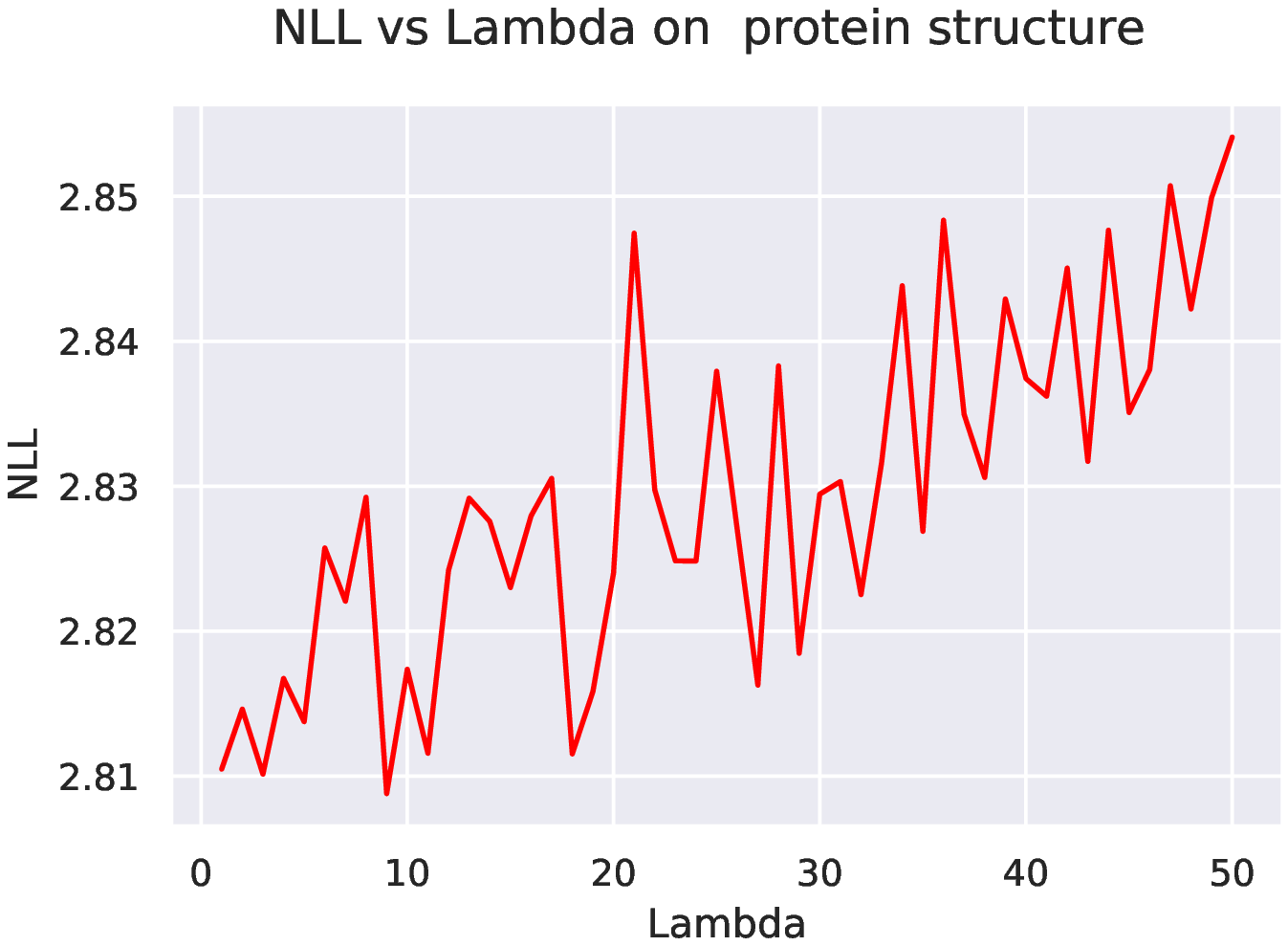}
    \includegraphics[width = 0.3 \linewidth]{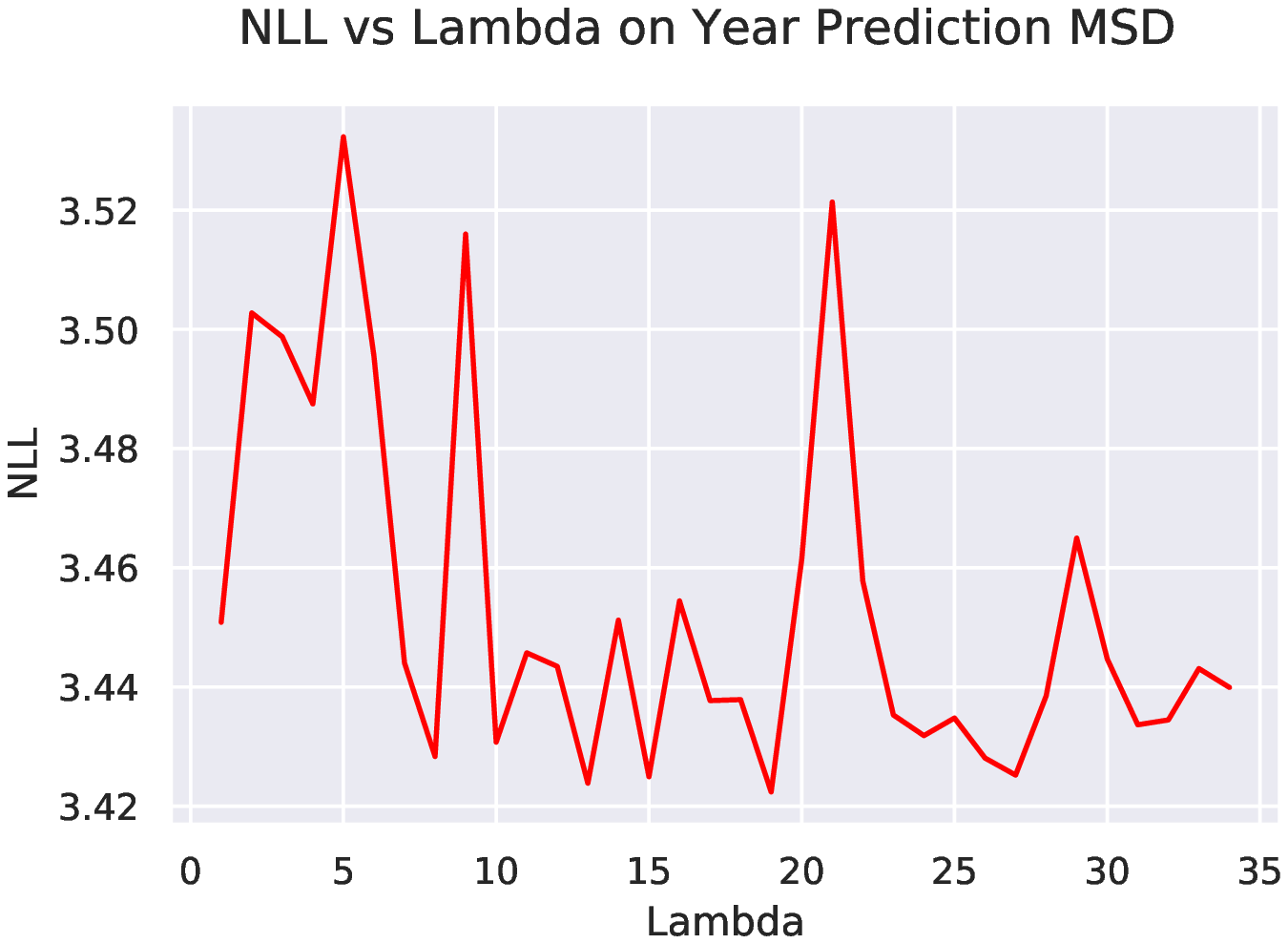}
    \caption{We can see that Calibration error reduces gradually as we increase $\lambda$ on Concrete Strength , Protein Structure , Year Prediction MSD dataset}
    \label{fig:fig2}

\end{figure}

\hyperref[fig:fig2]{Figure2} shows how calibration error, RMSE, and NLL changes as we vary $\lambda$ . We emphasize that there is no significant change in NLL/RMSE. Note that, while reporting results in the tables, we always set $\lambda = 20$.

\textbf{Post-hoc Recalibration }

\hyperref[tab:third]{Table 3} and \hyperref[tab:fourth]{Table 4} compare results when isotonic calibration is done . With Dropout-VI as base model  in $5/10$ cases, post-processing worsens calibration error.  These are instances where over-fitting of isotonic regression can be manifested.  Isotonic calibration works well on large datsets like Kin8nm Protein Structure, Year Prediction MSD; one possible justification is that there is plenty of data to recalibrate in these cases. Similarly coming to Deep Ensembles we can see same phenomenon, that post-hoc processing can increase calibration error . Here it is even more worse cause $7/10$ cases increases calibration error. The amount of miscalibration is much more in case of Deep Ensembles when compared to MC dropout Model. However, note that, very large datasets like Year Prediction MSD isotonic calibration does perform well just like anticipated. Another thing to be noted is that the amount of increase in miscalibration is smaller when the model is trained with Quantile Regularization.

\begin{table}[]
    \centering
   
     \label{tab:third}
    \scalebox{0.6}{\begin{tabular}{ccccc}
        \toprule
        
        Dataset & \multicolumn{4}{c}{HeteroScedastic MC Dropout model}                                                                          \\ \cmidrule{2-5}
        
        & \multicolumn{2}{c}{Calib Error(\%)} & \multicolumn{2}{c}{Iso Recalib Error(\%)}   \\ \cmidrule{2-5} 
        
        & base             & base+iso      & QR             & QR + iso       \\
        \midrule
        Air Foil$^*$  &  25.92 $\pm$ 5.40  & 41.32 $\pm$ 6.49 & \B 19.00 $\pm$ 4.80  &  \B 23.04 $\pm$ 2.42   \\
        
       Boston Housing$^*$      & 44.99 $\pm$ 4.41 & 56.78 $\pm$ 7.25 &  \B 42.59 $\pm$ 5.30  & \B 54.42 $\pm$ 2.99 \\

        Concrete Strength      & 58.75 $\pm$ 8.31 & 56.32 $\pm$ 6.20 &\B 35.42 $\pm$ 6.32 &  \B 30.21 $\pm$ 8.40  \\

         Fish Toxicity$^*$      &4.22 $\pm$ 1.56 &  6.09 $\pm$ 0.56 & \B 3.94 $\pm$ 0.78 & \B 3.48 $\pm$ 0.34   \\
         
          Kin8nm       &  12.37 $\pm$ 0.61 &  0.46 $\pm$ 0.11 & \B 11.60 $\pm$ 1.12 &  \B 0.37 $\pm$ 0.06      \\ 
          
           Protein  Structure      & 5.49 $\pm$ 0.59  &  0.10 $\pm$ 0.01 & \B 3.79 $\pm$ 0.82 &  \B 0.09 $\pm$ 0.00     \\

            Red Wine$^*$     & 8.86 $\pm$ 1.11    &  17.42 $\pm$ 1.29 &  \B 6.78 $\pm$ 0.99 & \B 8.38 $\pm$ 0.60       \\

            White Wine$^*$     & 9.69 $\pm$ 1.96 & 17.58 $\pm$ 1.28 & \B 7.63 $\pm$ 1.82  &  \B 10.98 $\pm$ 0.33   \\ 
            
            Yacht  Hydrodynamics      & 55.21 $\pm$ 9.38   & 23.96 $\pm$ 5.92 &  \B 40.08 $\pm$ 8.85 & \B 17.51 $\pm$ 6.53    \\ 
            
            Year Prediction MSD      & 8.52 $\pm$ 2.42  &  0.25 $\pm$ 0.12 &   \B 3.89 $\pm$ 3.56 & \B 0.14 $\pm$ 0.07     \\ 
        
        \bottomrule
    \end{tabular}}
     \caption{Base Model is Dropout-VI model. $*$ indicates the datasets for which isotonic recalibration \emph{increases} the calibration error, especially on smaller datasets. }
\end{table}

\begin{table}[]
    \centering
   
     \label{tab:fourth}
    \scalebox{0.6}{\begin{tabular}{ccccc}
        \toprule
        
        Dataset & \multicolumn{4}{c}{Deep Ensembles}                                                                          \\ \cmidrule{2-5}
        
        & \multicolumn{2}{c}{Calib Error(\%)} & \multicolumn{2}{c}{Iso Recalib Error(\%)}   \\ \cmidrule{2-5} 
        
        & base             & base+iso      & QR             & QR + iso       \\
        \midrule
        Air Foil$^{*}$  & 45.04 $\pm$ 0.86  & 79.00 $\pm$ 5.24  & \B 30.63 $\pm$ 3.67 &  \B 42.07 $\pm$ 2.21    \\
        
       Boston Housing$^{*}$      & 8.19 $\pm$ 2.07 & 33.99 $\pm$ 2.13  &  \B 5.89 $\pm$ 1.50 & \B 19.00 $\pm$ 1.50 \\

        Concrete Strength$^{*}$      & 81.34 $\pm$ 4.61   & 121.60 $\pm$ 8.60  & \B 65.48 $\pm$ 3.27 & \B 69.42 $\pm$ 5.93   \\
        
         Fish Toxicity$^{*}$      &5.35 $\pm$  0.65  & 19.44 $\pm$ 0.33 & \B 4.39 $\pm$ 1.23 & \B 6.89 $\pm$ 0.37     \\
         
          Kin8nm$^{*}$       & \B 1.22 $\pm$ 0.30  & 19.07 $\pm$ 0.41 & 1.54 $\pm$ 0.43& \B 6.23 $\pm$ 0.28    \\

           Protein  Structure      & 5.09 $\pm$ 0.39  & \B 0.09 $\pm$ 0.01 & \B 3.07 $\pm$ 0.20 & \B 0.09 $\pm$  0.00      \\ 
           
            Red Wine$^{*}$      &9.56 $\pm$ 1.21  &  35.39 $\pm$ 1.02 &   \B 7.16 $\pm$ 1.15 &  \B 13.95 $\pm$ 0.59       \\

            White Wine$^{*}$      & 8.42 $\pm$ 0.55  & 25.70 $\pm$ 0.50   & \B 8.06 $\pm$ 0.86 & \B 13.22 $\pm$ 0.54   \\ 
            Yacht  Hydrodynamics      &84.38 $\pm$ 4.58 & \B  50.24 $\pm$ 19.04  &  \B 54.23 $\pm$ 5.65  & 71.67 $\pm$ 8.65    \\ 
            
            Year Prediction MSD      & 6.57 $\pm$ 1.39 & 0.07 $\pm$ 0.02 & \B 2.41 $\pm$ 1.22  & \B0.05 $\pm$ 0.01   \\ 
        
        \bottomrule
    \end{tabular}}
     \caption{Base Model is Dropout-VI model. $*$ indicates the datasets for which isotonic recalibration \emph{increases} the calibration error, especially on smaller datasets. }
\end{table}

\section{Conclusion and Future work}
\label{sec:Future}

Although there is significant empirical evidence that calibrated models produce more reliable uncertainty estimates and generalize well, there is relatively less theoretical understanding as to why calibrated models are superior. Properties of calibrated classification models were studied in~\cite{cohen2004properties}; however, an in-depth analysis of the properties of calibrated regression models is currently lacking. Also, as mentioned in~\cite{2019arXiv190506023S}, Quantile Calibration is based on marginal probabilities. A more stronger notion is Distributional Calibration. As interesting avenue of future work will be to design trainable loss functions for the notion of Distributional Calibration.

\newpage

\section*{Appendix}
\label{sec:Appendix}

\renewcommand{\thesubsection}{\Alph{subsection}}
\subsection*{A1 : Proofs}

~

Proof of proposition 1

\begin{proof}
 To show that $R \circ F$ is quantile calibrated. we have to show that $\mathbb{P}[(R \circ F)[X][Y] \leq p] = p $ $\forall p \in [0,1]$ since we are assuming that $R(p)$ is invertible function, which gives us that it is surjective. So, an equivalent way of showing this is that $\mathbb{P}[(R \circ F)[X][Y] \leq R(p)] = R(p) $ $\forall p \in [0,1]$ 
 
  \begin{align*}
     &= \mathbb{P}\bigg[(R \circ F)[X][Y] \leq R(p)\bigg] \\
     &= \mathbb{P}\bigg[ R^{-1} \big((R \circ F)[X][Y] \big) \leq R^{-1}\big(R(p)\big)\bigg] &&  R^{-1} \text{ is strictly increasing}\\ % R^{-1} \text{ is strictly increasing}   
    &= \mathbb{P}\bigg[  (F[X])[Y]  \leq p \bigg]  \\
    &= R(p) &&  \text{By definition} \\ 
  \end{align*}
\end{proof}

Proof of proposition 2

\begin{proof}

 Note that the expectation and survival function of $ \text{Uniform}[0,1]$ are $\mathbb{E}[T] = 0.5$ and $\overline{G}_{T}(x) = 1-x$,  respectively. Let $c(x) =  \int \ln(1-x) = -(1-x) \ln(1-x) + 1-x$ 
and note that $\lim_{x \rightarrow 1} c(x) = 0$ and $\lim_{x \rightarrow 0} c(x) =1$
 
  \begin{align*}
     &= \int_0^{1} \overline{F}_{S}(x) \ln \frac{\overline{F}_{S}(x)}{\overline{G}_{T}(x)} dx - \mathbb{E}[S] + \mathbb{E}[T]  \\
     &=\int_0^{1} \overline{F}_{S}(x) \ln \frac{\overline{F}_{S}(x)}{\ln(1-x)} dx - \mathbb{E}[S] + 1/2 \\ 
    &= \int_0^1 \overline{F}_{S}(x) \ln \overline{F}_{S}(x) dx - \int_0^1 \overline{F}_{S}(x) \ln(1-x) dx - \mathbb{E}[S] + 0.5\\ 
    &=  -\epsilon(F_{S}) - \Big[ c(x) \overline{F}_{S}(x) \Big]_{0}^1 + \int_0^1 c(x)   [ -f(x)] dx - \mathbb{E}[S] + 0.5  \\ 
    &= -\epsilon(F_{S}) + 1 +  \mathbb{E}[(1-S) \ln (1-S) + (S-1)] - \mathbb{E}[S] + 0.5 \\
    &= - \epsilon(F_{S}) + \mathbb{E}[(1-S) \ln (1-S)] + 0.5
  \end{align*}
\end{proof}

% \begin{claim}

% Let $X \sim F$ , with support $[0,1]$ then

% \begin{ceqn}
% \begin{align*}
%   \int_0^1 \overline{F}(x) \ln(1-x) dx  = \text{E}[- (1-X) \ln (1-X) + (1-X) ] 
% \end{align*}
% \end{ceqn}

% \end{claim}

Proof of proposition 3
~

\begin{proof}
\cite{rao2004cumulative} show that $\epsilon(F_S^{n}) \rightarrow \epsilon(F_S)$ where $F_S^{n}$ is empirical CDF from n samples.let $x_{(0)} = 0$ with $F_S^n(x) = \sum_{i=0}^{N-1} \frac{i}{N} \text{I}[s_{(i)} \leq s < s_{(i+1)}]$

We have that 
 $F_S^n(x) = \sum_{i=0}^{n-1}  \frac{n-i}{n} \, \text{I}[s_{(i)} \leq s < s_{(i+1)}], $ $\epsilon(F_S^n) =   \sum_{i=1}^{n-1} \frac{n-i}{n} \Big (\ln \frac{n-i}{n}  \Big)  s_{(i+1)} - s_{(i)}.$ the expectation  $\mathbb{E}[ (1-S) \ln(1-S)]$ can be replaced by sample mean $ \frac{1}{n}\sum_{i=1}^n (1- s_i) \ln (1-s_i).$ so, overall, we have a consistent estimator.

\end{proof}

\bibliography{qr.bib}

\end{document}